# Hardware design for binarization and thinning of fingerprint images

Farshad Kheiri, Shadrokh Samavi, Nader Karimi,

**Abstract**: Two critical steps in fingerprint recognition are binarization and thinning of the image. The need for real time processing motivates us to select local adaptive thresholding approach for the binarization step. We introduce a new hardware for this purpose based on pipeline architecture. We propose a formula for selecting an optimal block size for the thresholding purpose. To decrease minutiae false detection, the binarized image is dilated. We also present in this paper a new pipeline structure for implementing the thinning algorithm

## 1 Introduction

In today's life style, security is one of the most important concerns. Technology could provide us with this purpose. Perhaps the first step is to identify a person. Identification cards or simple identification numbers are two examples of this system. By identifying a person more precisely, the system's security can improve.

One of the methods to identify a person is biometrics. It is defined as a science which studies human's behaviors and physical characteristics, to identify him/her [1]. This identification is done by recognition of face, hand, voice, retina, iris and fingerprint. Uniqueness and permanency are two main concerns to select a biometric characteristic. Although iris recognition is one of the most precise methods, it is not acceptable by all the people as a non-invasive method. Since eyes are scanned by infrared, people are afraid of hurting their eyes [1]. Fingerprint as a permanence and uniqueness characteristic, is acceptable by people as a non-invasive routine. Based on what is reported by International Biometric Group in 2002, as shown in Fig. 1, fingerprint recognition is used in 52.1 percents of biometric systems [2].

Fingerprint recognition was first employed by Scotland Yard in 1901 as an identification system for the first time [3]. They used Henry-Galton system. This system identifies five types of fingerprint which are shown in Fig. 2. The categories that are shown in Fig. 2 are arches, tented arches, left loops, right loops, and whorls.

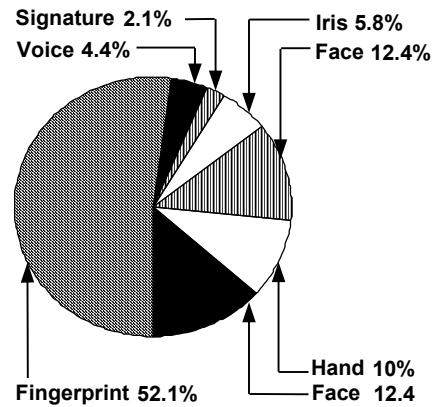

Fig. 1. Biometric market report in 2002 [2].

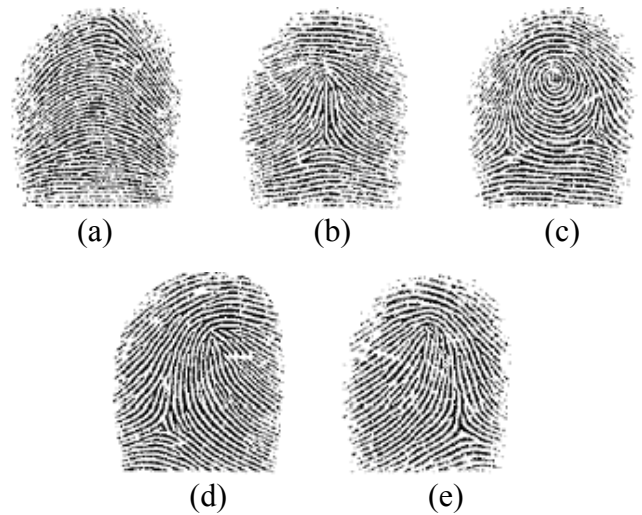

Fig.2: Five categories of fingerprints [3] (a) arch, (b) tented arch, (c) whorl, (d) right loop, (e) left loop.

Farshad Kheiri is with the Department of Electrical and Computer Engineering, The University of Alabama, Huntsville, Al, 35899 USA.
S. Samavi and N. Karimi are with the Department of Electrical and Computer Engineering, Isfahan University of Technology, Isfahan, 84156, Iran.

Later Galton introduced a more precise system, based on minutiae points. Among minutiae points, two characteristics are considered more than others. These minutiae points are ridge ending and ridge bifurcation [4]. They are shown in Fig.3.

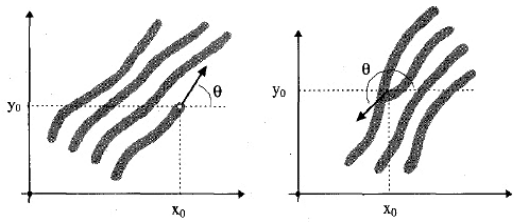

Fig. 3. Most common minutiae features (a) Ridge endings, and (b) ridge bifurcations [4].

In AFIS (Automated Fingerprint Identification Systems), fingerprints are first classified in five mentioned types, and then recognition is performed based on the minutiae locations.

Extracting minutiae points and matching them in two fingerprints are two main tasks in the recognition process. To extract minutiae, a fingerprint image should have 40 to 100 minutiae [3]. To match two fingerprints, between 10 to 16 minutiae should match. The variation is related to different human races [3].

For an input image the following pre-processing stages which are shown in Fig. 4, can be used before minutiae extraction.

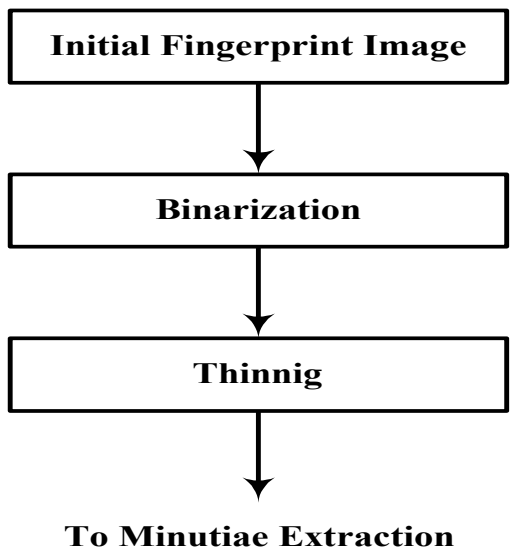

Fig. 4: The common preprocessing stages used before minutiae extraction [9].

In binarization, a binary image will be obtained from a gray-scale image [5]. Then the binarized image could be thinned to produce lines that are almost one pixel width [6]. By thinning, minutiae extraction will be easier than before [4]. It is obvious that applying a poor binarization method will affect all other stages adversely. Therefore it can be a critical step in fingerprint recognition. To improve binarization and reduce false minutiae detection, we introduced an extra stage, dilation, between two mentioned stages [7].

For fingerprint image processing, image production could be performed by line-scan procedure and on the other hand the need for a real-time process results in applying a pipeline structure [8]. In this paper a new approach based on pipeline architecture is presented. This approach consists of thresholding, dilation and thinning. In part 2, binarization is explained in more details. Thinning algorithm is studied in part 3. In part 4, the need for dilation is expressed. The suggested pipeline structure is introduced at part 4 and then complexity analysis is discussed in part 5. Finally conclusions come in part 6.

## 2 Binarization

For an input image, some processing stages should be used before extracting minutiae. One of these stages is binarization. In this stage the gray-scale image converts into a binary image. A binary image can be processed better than a grayscale image [3]. As mentioned in [4], there are three main approaches in binarization:
1) Global Approaches,
2) Neighborhood-Based Approaches
3) Filter-Based Approaches.

In another point of view [9], approaches can be divided into two main categories:
1) Software routines
2) Hardware methods

Software approaches are computationally expensive, time-consuming, highly flexible, and non-real time. Filter-based approaches belong to this group. The other group which is of our interest can be implemented in real time systems. This group can be classified into global and local approaches. Global approach does not fulfill the requirements. As it is explained in [4], it can produce unfavorable results. These results can be the side effects of non- homogenous press of finger on the scanner. On the other hand, hardware implementation for this approach will be architecturally expensive [9].

Since we have limitation in the area of speed of processing, it seems that among the above approaches, local adaptive thresholding is a good choice. It has fast speed, low complexity, fair quality and good robustness [4].

The basic idea in thresholding is to select a threshold (T) to extract an object or several objects with the same value from background [10]. We limit our discussion to one-level thresholding for gray-scale images. Equation (1) can be used to binarize the grayscale images:

$$g(x,y) = \begin{cases} 1 & \text{if } f(x,y) > T \\ 0 & \text{if } f(x,y) \leq T \end{cases} \quad (1)$$

where f(x,y) is the value of a pixel in the grayscale image and g(x,y) is the binarized image. In adaptive thresholding, the image is divided into series of blocks. A threshold will be defined for each block. All pixels in the block will compare with this threshold. In this approach, we encounter with two main problems:
 a) What is the optimal size of a block?
 b) How is the threshold calculated for each block?

As is mentioned in [3], it has been empirically proven that blocks with size 16×16 are the best fixed size for binarization of fingerprint images. To prove this subject, it is known from [10], to select a good threshold, the histogram peaks should be separated by deep valleys.

By inspection, it seems that variance is a good factor to select a fixed size block. Firstly, each image is divided into number of equal blocks. Next, the threshold value is calculated inside each block. Thirdly, the mean gray scale value of all blocks is calculated. Then variance will be defined from [11].

$$\sigma^2 = 1/n \sum_{1}^{n-1} (T_i - \mu)^2 \quad (2)$$

$T_i$ is the threshold value of each block, μ is the mean of all blocks, n is the number of blocks. To incorporate the value of block size into the formula, we divide $\sigma^2$ by the forth root of block size. This formula satisfies the results found empirically. We, therefore, define the following factor for an N×N block [9]:

$$Block\ Factor = \sigma^2 \times \sqrt[4]{N} \quad (3)$$

We calculate this factor for various N's of size 4, 16, 64, and 256. These block sizes are for a 512×512 pixels image. We excluded 2, 8, 32, 128, and other block sizes since they did not produce good results. This factor results in 16×16 as being the choice. Results of applying different block sizes to an image are shown in Fig. 5. By inspection, it seems that 16×16 is the best size. The mentioned factor produced good results for a large number of images.

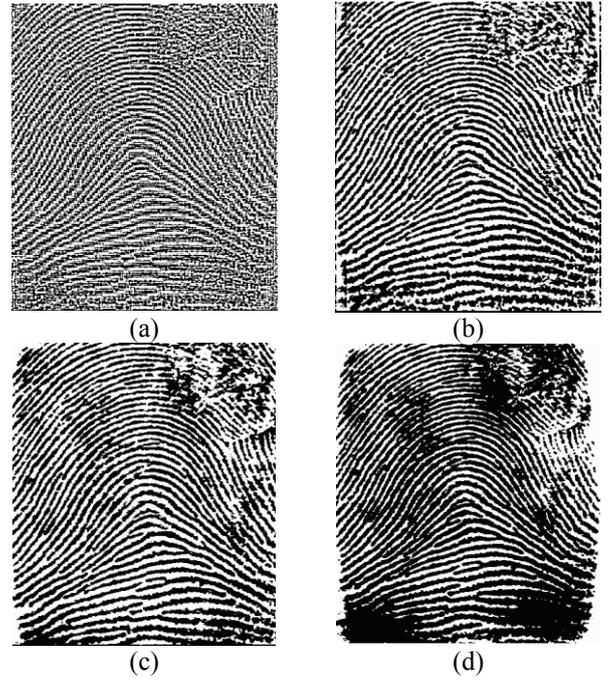

Fig. 5: Binarized image with block Size (a) 4×4 (b) 16×16 (c) 64×64 (d) 256×256 [9].

To select an optimal thresholding value, we should minimize the erroneous classification of ridges as valleys or vice versa. This is calculated in equation (4). By exploring the histogram of a typical fingerprint in Fig. 6, and the resemblance between this histogram and the one presented in [10], the ridges and valleys can be replaced by object and background respectively. Using the same argument as in [10] we can come up with the following threshold value:

$$T = (1/2)(\mu_1 + \mu_2) + (\sigma^2/(\mu_1 - \mu_2)) \ln(P_1/P_2) \quad (4)$$

where $\mu_1$ and $\mu_2$ are the mean values of the object and that of the background, $\sigma^2$ is the variance of the pixels. $P_1$ and $P_2$ are the probabilities of occurrence of the two classes of pixels. It is acceptable to assume $P_1 = P_2$ for fingerprint images if the region of interest in an image only covers the fingerprint image, not the surrounding background. This is valid assumption since the region of interest contains the same number of valley pixels as there are ridge pixels. Hence, the above equation can be simplified as:

$$T = (\mu_1 + \mu_2)/2 \quad (5)$$

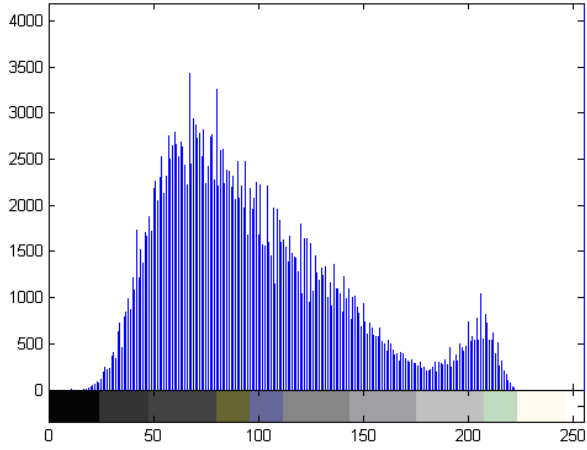

Fig. 6: Histogram of a typical fingerprint [4].

Another method of local adaptive thresholding is to use overlapped blocks [12]. We used mean square signal to noise ratio (SNR$_{ms}$), as described in equations (6), to find out which block size and how many overlapped pixels would produce better results. In other trials, we calculated root-mean-square-error ($e_{rms}$) and correlation with Otsu's routine. Furthermore, we considered Otsu's thresholding method [13] as a standard method. The Otsu's method is a viable software approach but is not an attractive one for hardware implementation. We compared our method with that of Otsu's. Using 16×16 blocks with one overlapped pixel between adjacent blocks produced the better results than 16×16 non-overlap blocks. In (6) and (7) variables G and F are respectively the pixels of the image processed by the Otsu's method and ours.

$$SNR_m = (\sum_{x=0}^{M-1}\sum_{y=0}^{N-1} G(x,y)^2) / (\sum_{x=0}^{M-1}\sum_{y=0}^{N-1} (G(x,y) - F(x,y))^2) \quad (6)$$

$$e_{rms} = (1/MN \sum_{x=0}^{M-1}\sum_{y=0}^{N-1} (G(x,y) - F(x,y))^2)^{1/2} \quad (7)$$

The results of the above equations for a set of sample images are calculated in table 1.

Table 1. Accuracy of the thresholding scheme

| Block size | Overlap pixels | SNR$_m$ | e$_{rms}$ |
|---|---|---|---|
| 16×16 | 0 | | |
| 16×16 | 1 | | |

The local adaptive thresholding is performed on an image in Fig. 7. As it can be seen some minutiae points have been damaged. Hence, the binarized image needs an extra step to enhance the image. This stage is the thinning process.

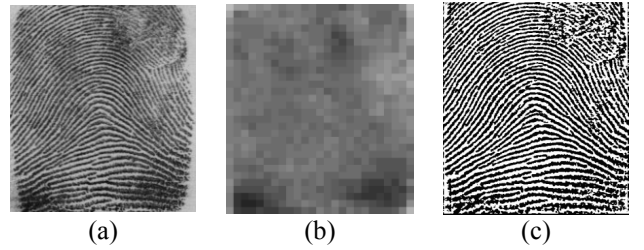

(a)      (b)      (c)

Fig. 7: An example of local adaptive thresholding with block size 16×16, (a) fingerprint (b) block threshold value, (c) binarized image [4].

## 3 Thinning process

From what is presented in [6], thinning is defined as a procedure to transform a digital binary pattern to a connected skeleton of unit width. Two basic implementations available for this approach are *sequential* and *parallel* methods.

In the sequential method, based on the results that are obtained, all pixels are examined and changed. The parallel method works on all pixels simultaneously. As discussed in [6] parallel thinning is substantially faster on pipeline computers. It is also mentioned in [14] that in all thinning algorithms only values of the closest neighbors are needed to remove a pixel. Therefore a window could be moved on the image in order to examine the center pixel based on its neighboring pixels.

The need for thinning process is discussed extensively in [15] and [16]. From what is indicated in [3], thinning procedure increases the speed of minutiae extraction.

Out of all thinning algorithms, what are presented in [17] and [18], are selected for our purpose. Simplicity and fixed-size window (3×3) of this algorithm motivated us to choose this method. There are algorithms that use different windows [6]. Of course we modified the algorithm to fulfill the needs of our application.

To implement the thinning algorithm, based on what is presented in [17], four conditions in each sub-iteration should be provided. There are two common conditions in each sub-iteration which are:

$$a) \; 3 \leq B(P_1) \leq 6 \quad (8)$$
$$b) \; A(P_1) = 1 \quad (9)$$

The number of 01 patterns in the ordered set $P_2, P_3, \ldots, P_8, P_9$ is $A(P_1)$ and the sum of $P_2, P_3, \ldots, P_8, P_9$ is $B(P_1)$. The locations of $P_2, P_3, \ldots, P_8, P_9$ in relation to $P_1$ are shown in Table 2.

There are 512 possible combinations that these one bit pixels could have. $P_1$ should be 1, therefore the number of cases reduce to 256. Based on the first condition, it is concluded that the number of

1's in $P_2$ to $P_9$ should be 3, 4, 5 or 6. If we consider the neighboring pixels of $P_1$ as forming a ring then the next condition indicates that the 1's in this ring should follow each other. It means that there should be no gap between 1's in the string. The first member of this group of 1's can be located in eight different positions.

Table 2. Position of pixels in window [17].

| $P_9$ (i-1,j-1) | $P_2$ (i-1,j) | $P_3$ (i-1,j+1) |
|---|---|---|
| $P_8$ (i,j-1) | $P_1$ (i,j) | $P_4$ (i,j+1) |
| $P_7$ (i+1,j-1) | $P_6$ (i+1,j) | $P_5$ (i+1,j+1) |

The number of possible combinations will reduce to 32 terms; there are 8 permutations for each sum [9]. The other two conditions that should be satisfied in each sub-iteration are as follows.

$c : P_2 \ \& \ P_4 \ \& \ P_6 = 0$ (10)

$d : P_4 \ \& \ P_6 \ \& \ P_8 = 0$ (11)

$c' : P_2 \ \& \ P_4 \ \& \ P_8 = 0$ (12)

$d' : P_2 \ \& \ P_6 \ \& \ P_8 = 0$ (13)

Therefore, there are two sub-iterations. Sub-iteration *I* should satisfy conditions a, b, c and d. Then sub-iteration *II* should satisfy conditions a, b, c', and d'. Conditions c and d cause combinations shown in table 2 to occur.

Table 2. Combinations omitted by conditions c and d [4].

| $P_2$ | $P_3$ | $P_4$ | $P_5$ | $P_6$ | $P_7$ | $P_8$ | $P_9$ |
|---|---|---|---|---|---|---|---|
| 1 | 1 | 1 | 1 | 1 | 0 | 0 | 0 |
| 0 | 0 | 1 | 1 | 1 | 1 | 1 | 0 |
| 1 | 1 | 1 | 1 | 1 | 1 | 0 | 0 |
| 0 | 1 | 1 | 1 | 1 | 1 | 1 | 0 |
| 0 | 0 | 1 | 1 | 1 | 1 | 1 | 1 |
| 1 | 1 | 1 | 1 | 1 | 0 | 0 | 1 |

No combinations should omit from 3 or 4 ones, because from $P_2$ to $P_6$ and from $P_4$ to $P_8$ there are 5 gaps, therefore these two conditions will be satisfied for all 3 or 4 connected-ones. Finally 26 combinations will remain. Each combination is considered as a minterm and its logic function is realized. The *thinning processor circuit* (TPC) is implemented based on these 26 combinations [9].

The number of repetition of the iterations should be discussed as another topic. It is indicated in [19] that the value of the spatial frequency of the ridges and valleys in a local neighborhood lies in a certain range for 500 dpi image. This range is [1/3, 1/25]. Half of the inverse of this value determines the thickness of ridges. Therefore, at most, six iterations are needed to be performed. Each time two pixels are omitted after applying the thinning algorithm. This has been proved experimentally. For a group of fingerprint images, a number of iterations are applied and then the initial binarized image is subtracted from this image for a region of interest. The total number of changed pixels for each iteration is then normalized. Fig. 8 shows that after six iterations the thinning process is exhausted [9].

It can be observed that after applying iterations six times, variations in values of pixels will be negligible. For thinner ridges this can take place sooner [9].

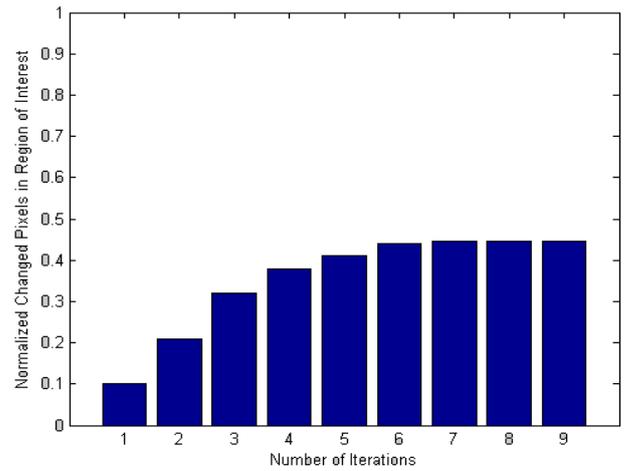

Fig. 8. Change in number of pixels vs. number of iterations [9].

## 4 Dilation

Whereas binarization is not ideal, after thresholding there will be some blank gaps, known as holes, and disconnection in ridges. In the next stage, holes may be changed to bifurcation points and disconnections may be converted to ridge endings. To compensate this false minutiae extraction, dilation can play a main role. This can be seen in Fig. 9.

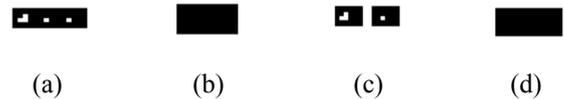

(a)     (b)     (c)     (d)

Fig. 9. (a) image with holes, (b) dilation results, (c) image with holes and disconnection, (d) dilation results [12].

As suggested in [10], two morphological elements are applied to avoid noise in binary image for fingerprint. In opening, erosion reduces or deletes the noise of background, while it enlarges the noise of inner boundaries. Dilation decreases this noise. Whereas we applied a morphological thinning process, this process will omit the noise of background. Therefore it doesn't need erosion in opening. As mentioned in [10], closing is applied to avoid disconnections. This can be performed by dilation too. It seems one dilation stage can lonely maintain these two duties. Dilation has applied to a sample fingerprint image in Fig. 10.

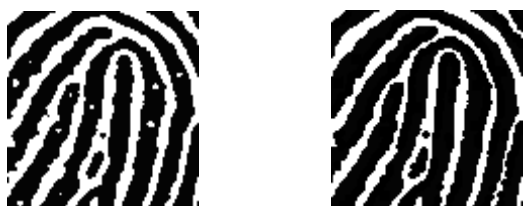

(a) Binanarized image  (b) Dilated image
Fig.10. Dilation of a sample image.

## 5 pipeline structure

This architecture has been designed based on some pre-assumptions:
1) Each line of image has 512 pixels,
2) Each pixel's value is stored in a 8-bit register,
3) Input and output bus are 32-bit,
4) Structure has a main clock pulse and all other pulses are made of it,
5) Frequency of pipeline is determined by the most delay part.

The first and last stage of the pipeline is shown in Fig. 11(a). The first stage receives the image through a 32-bit bus and distributes it on 512 registers. In each clock pulse, four registers are loaded. The first 8-bit of the input bus is connected to the first register, the second 8-bit of the input bus is connected to the second register, the third 8-bit of the input bus is connected to the third register and finally the fourth 8-bit of the input bus is connected to the fourth register. On the other hand, the first register is connected to the fifth, ninth, thirteenth … registers through another bus. The same connections exist for the second, third and fourth register. A decoder is applied to select which register should be loaded in each stage. There are 512 registers and in each clock pulse, four of them will be loaded, therefore 128 clock pulses are necessary to load one line of the image. These clock pulses activate the load pins of registers through a 128-output decoder. A 7-bit counter feeds this decoder. Main clock enters to the 7-bit counter. It is obvious that after 128 clocks, the first line should go to the next stage of the pipeline; therefore, the main clock divides into 128 to produce the clock of pipeline. The last stage sends the image out of the architecture through a 32-bit bus. Whereas the image is binarized at the end, the pixels of the output image have one-bit; therefore, 512 bits exit in 16 clocks. There are 512 buffers in the output. The load pins of each 32 buffers are connected to each other and the outputs of a 16 output decoder enter to the load pins of each 32-buffer group. The main clock divided by 16, feeds this decoder. It has been shown in Fig. 11.b. To binarize a gray-scale image based on adaptive thresholding approach, firstly the image is divided into series of 16×16 blocks. Then the mean gray scale value of each block is calculated as the threshold, T, for that block. By comparing each pixel in a block with this mean value binarization is performed. Each block has one pixel overlap with its neighbors.

We need to add the pixels of a 16×16 blocks in a circuit. The number of inputs for such circuit is large. To reduce the number of inputs a pipeline is devised which loads only one row of pixels of block at each clock pulse. This means that sixteen 8-bit values are loaded at each clock pulse. Therefore, for every sixteen pixels there is a *mean value calculator unit* (MVCU).

To calculate the mean value of 16×16 block all pixels are added and the sum is divided by 256. In a pipeline structure to implement the MVCU a seventeen 8-bit inputs adder is needed. This adder is to take 16 pixels of each line of a block and add them with the results obtained from previous line which is another 8-bit input. This last input is a feedback from the low part of the output of the adder. To divide the total result into 256 an 8-bit right shift is needed. Fig. 12 shows the structure of a MVCU.

There are two adders in the circuit of Fig. 8. The adder with 17 inputs is more time demanding and requires special attention. Therefore, after applying different methods it was concluded that Dadda's method which uses Carry Save Adders (CSA) produces least delay [20]. Since there are seventeen 8-bit inputs, based on the Dadda's design an initial layer is required to reduce the number of inputs to thirteen. Four CSA's are employed in the first layer and then five layers of CSA are used to reduce the thirteen inputs to two outputs. The internal structure of Dadda's tree adder is shown in Fig. 13. As it can be seen in Fig. 13, five types of CSA are used in this design. All types are depicted in Fig. 14. At a final stage a two 4-bit carry look-ahead adder (CLA) is used to add the two final outputs, it is shown in Fig. 15.

The next 15 stages of the pipeline, which are shown in Fig. 16, perform the binarization process. Every row of the image that comes into the pipeline is divided among 34 MVCU's. By the time that 16 rows of the image are loaded into the pipeline each MVCU has calculated the mean gray scale value of a 16×16 block. The output of a MVCU is latched into an 8-bit register. This register is clocked once every 16 clock pulse of the regular pipeline stage. At this time the pixels of each of these 16 rows are compared with the corresponding latched outputs of MVCU's to produce a row of a binary image.

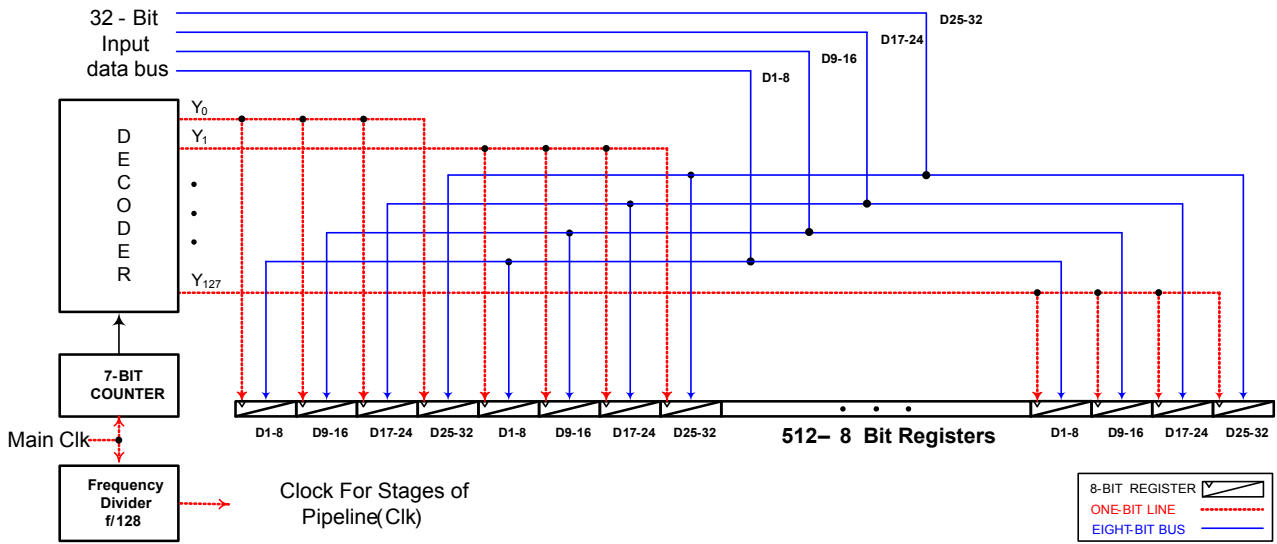

Fig. 11. First stage of the pipeline responsible for receiving the image through a 32-bit bus [4].

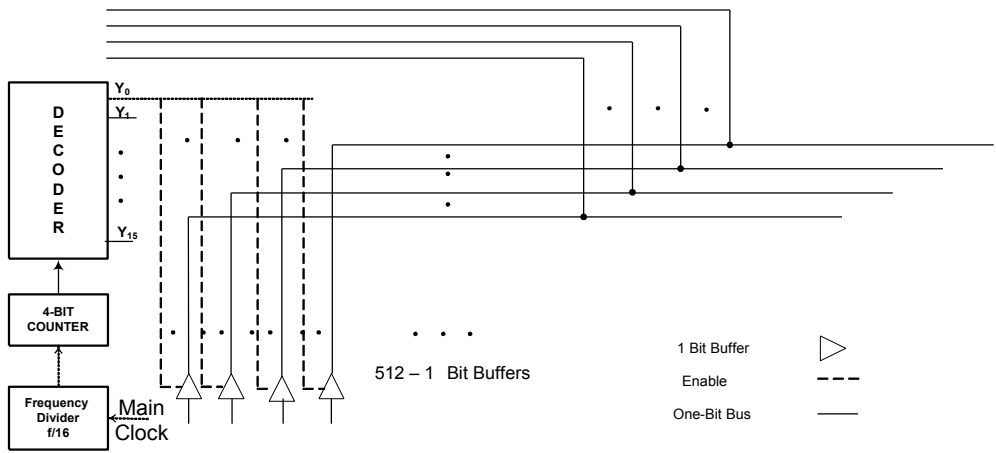

. Fig. 12. Last stage of pipeline through one-bit buffers.

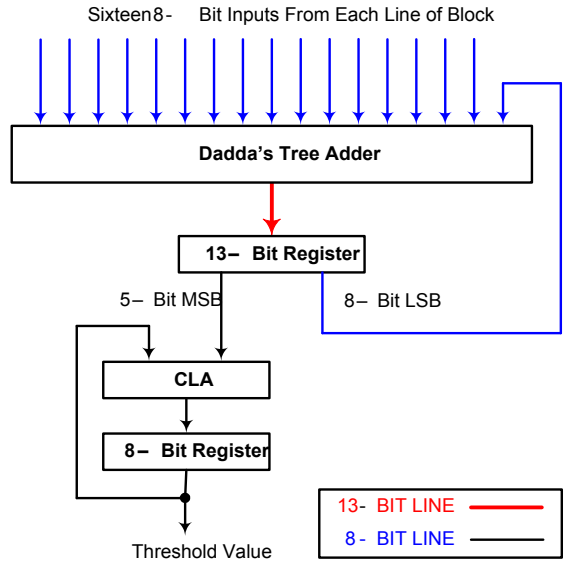

Fig. 12. Internal structure of a MVCU.

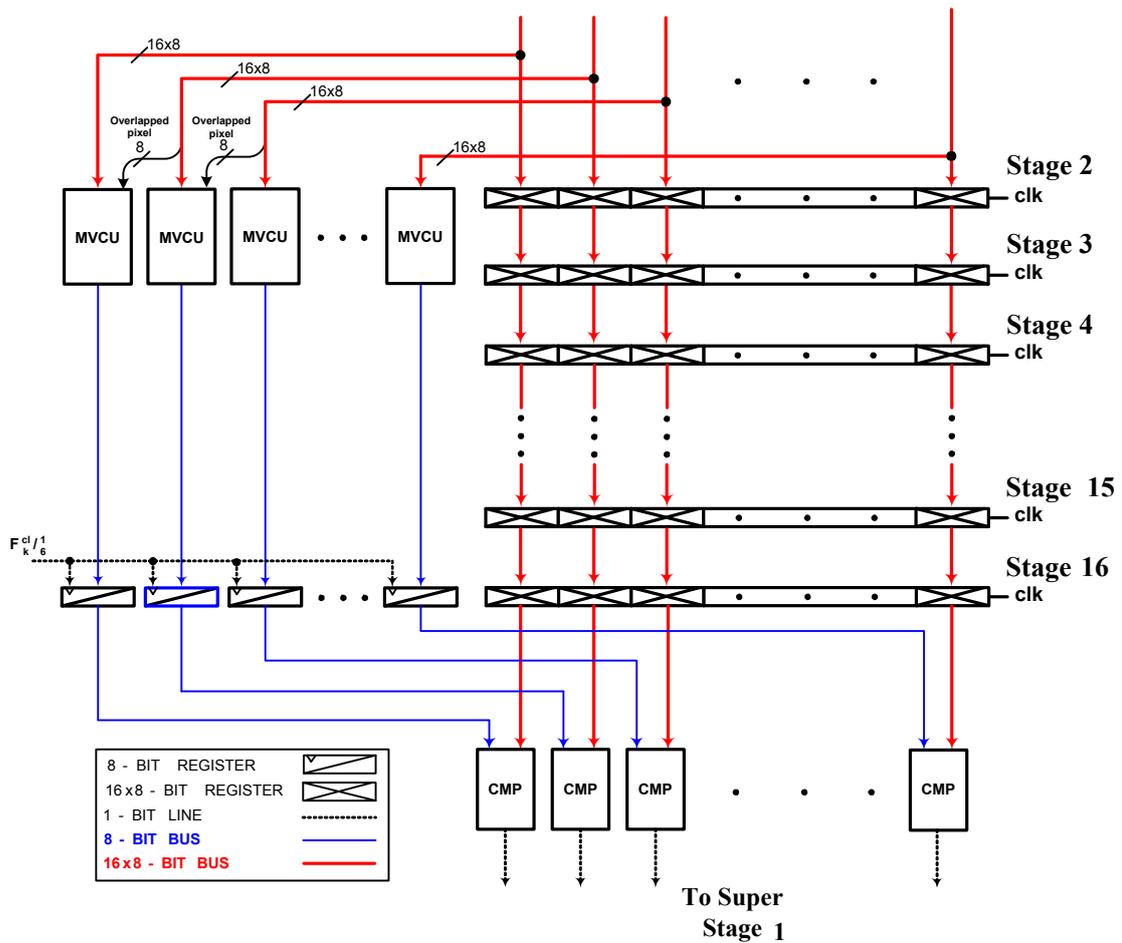

Fig. 16. Stages of the pipeline responsible for binarization.

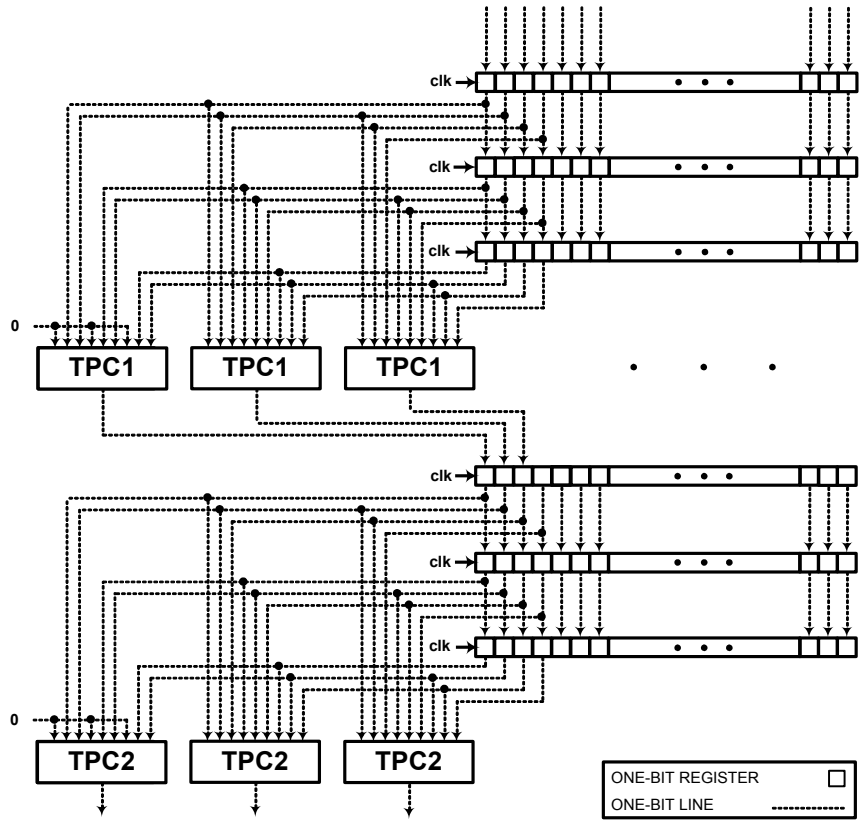

Fig. 17. The pipeline structure of thinning algorithm.

The hardware design that we propose for the thinning process has advantages over the methods presented in [6] and [21]. One of these advantages is the application of a fixed size window which results in easier implementation. Also in comparison with [22] our design is advantageous. We implement the modified thinning algorithm by changing the first condition mentioned in section 3. We also simplified the implementation of conditions, whereas in [22] a direct method is applied. Furthermore, in [22] the number of iterations that are required is not mentioned. Our method uses a fixed number of iterations.

Fig. 17 shows the hardware required for one iteration of the thinning process. The thinning mask that is used is a 3×3 window. Since two sets of conditions should be checked six pipeline stages are required to implement one iteration of the thinning algorithm. The binarized image is loaded into three stages. The nine pixels of the window are loaded into a *thinning processor circuit* (TPC1) which implements the first thinning condition. Then the outputs of TCP1's are loaded into the next three stages of the pipeline. The second set of conditions will be checked by a TCP2.

Based on results that are presented in Fig.9 these conditions should be examined six times. We refer to the six stages of Fig. 17 as one super stage. Therefore six super-stages are required to complete the thinning process.

Dilation can be implemented through an optimized architecture which is shown in Fig. 18, under the binarization stage. In fact, in this implementation, one pixel sets '1', if one of its neighbors in a 2×2 window is '1'.

## 5 Experimental results

The suggested structures were implemented using Virtex-II-Pro FPGAs from Xilinx Corporation. The simulation results show that the longest delay occurs in the MVCU and it equals to 6.29 nanoseconds. This caused the processing time for a 512*512 image to be about 6.84 milliseconds. The clock frequency on a xc2vp20 chip turned out to be about 79.4 MHz. The percentage of used slices on two kinds of Virtex-II-Pro FPGA are shown in table 3.

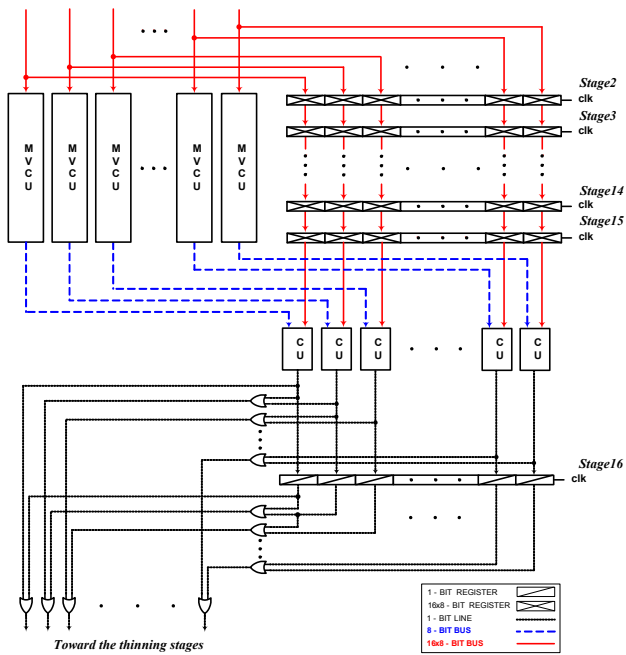

Fig. 18. Stages of the pipeline responsible for binarization and dilation processes.

Table 3. Implementation results

| FPGA type | Percentage of Employed CLB Slices |
|---|---|
| Xc2vp20 | 97.8% |
| Xc2vp100 | 20.6% |

Fig. 19(a) shows a typical fingerprint image that is processed by the proposed hardware. Fig. 19(b) depicts a binarized image produced by the binarization section of the pipeline. Fig. 19(c) is the dilated image and final thinned image that the proposed pipeline has produced, is shown in Fig. 19(d).

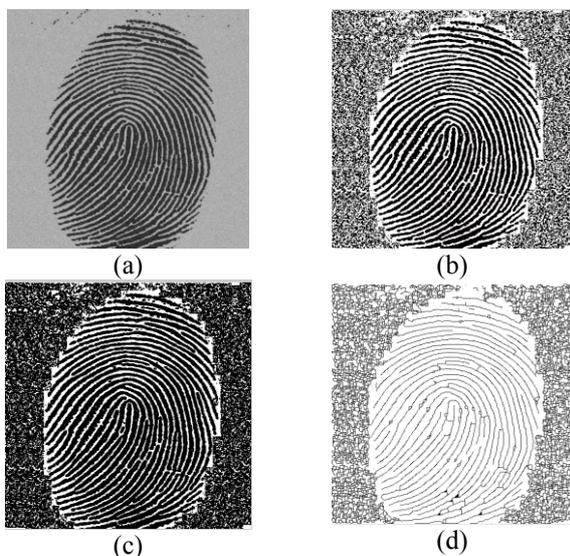

Fig. 19. (a) Sample fingerprint image, (b) binarization, (c) image after dilation, (d) result of thinning stage.

We compared our thinning architecture with Hsiao's method [22]. Although we used the same algorithm and implementation platform, Spartan-II xc2s100, their processing time was 70 milliseconds while ours was 10.36 milliseconds. It should be reminded that Hsiao used an external memory in his design, which requires extra hardware.

## 6 Conclusions

This paper presented an improved binarization, dilation and thinning algorithm. We offered an efficient architecture for implementing these algorithms. The proposed method, which is based on pipeline structures, is well suited for real time processing of fingerprint images when line scanners are used. We were able to achieve execution times of less than 7ms. Numerous images have been processed by the proposed hardware which proved its correct functionality and performance. Furthermore the structure is highly modular and expandable. Each part of the proposed hardware can be used independent of the other part. This is why we were able to compare the thinning part of our design with Hsiao's routine. Better hardware performance is expected if the suggested architecture were to be implemented on silicon.